\documentclass{article} %
\usepackage{iclr2017_conference,times}
\usepackage{hyperref}
\usepackage{url}

\usepackage{siunitx}
\usepackage{amsmath}
\usepackage{xcolor}
\usepackage{graphicx}
\usepackage[export]{adjustbox}

\usepackage{xspace}

\usepackage[utf8]{inputenc} %
\usepackage[T1]{fontenc}    %
\usepackage{hyperref}       %
\usepackage{url}            %
\usepackage{booktabs}       %
\usepackage{amsfonts}       %
\usepackage{nicefrac}       %
\usepackage{microtype}      %

\newcommand{\Wh}{\mathbf{W^h}}
\newcommand{\Wx}{\mathbf{W^x}}
\newcommand{\WW}{\mathbf{W}}
\newcommand{\bb}{\mathbf{b}}

\newcommand{\Wih}{\mathbf{W^{ih}}}
\newcommand{\Wix}{\mathbf{W^{ix}}}
\newcommand{\Wfh}{\mathbf{W^{fh}}}
\newcommand{\Wfx}{\mathbf{W^{fx}}}
\newcommand{\Woh}{\mathbf{W^{oh}}}
\newcommand{\Wox}{\mathbf{W^{ox}}}

\newcommand{\Wch}{\mathbf{W^{ch}}}
\newcommand{\Wcx}{\mathbf{W^{cx}}}
\newcommand{\Wgh}{\mathbf{W^{gh}}}
\newcommand{\Wgx}{\mathbf{W^{gx}}}
\newcommand{\Wrh}{\mathbf{W^{rh}}}
\newcommand{\Wrx}{\mathbf{W^{rx}}}
\newcommand{\Wuh}{\mathbf{W^{uh}}}
\newcommand{\Wux}{\mathbf{W^{ux}}}

\newcommand{\hht}{\mathbf{h}_t}
\newcommand{\xxt}{\mathbf{x}_t}
\newcommand{\cct}{\mathbf{c}_t}
\newcommand{\rrt}{\mathbf{r}_t}
\newcommand{\uut}{\mathbf{u}_t}
\newcommand{\ggt}{\mathbf{g}_t}

\newcommand{\iit}{\mathbf{i}_t}
\newcommand{\fft}{\mathbf{f}_t}
\newcommand{\oot}{\mathbf{o}_t}
\newcommand{\yyt}{\mathbf{y}_t}
\newcommand{\hhtm}{\mathbf{h}_{t-1}}

\newcommand{\bbc}{\mathbf{b^c}}
\newcommand{\bbg}{\mathbf{b^g}}
\newcommand{\bbr}{\mathbf{b^r}}
\newcommand{\bbu}{\mathbf{b^u}}
\newcommand{\bbh}{\mathbf{b^h}}
\newcommand{\bbi}{\mathbf{b^i}}
\newcommand{\bbf}{\mathbf{b^f}}
\newcommand{\bbo}{\mathbf{b^o}}

\newcommand{\hp}{HP\xspace}
\newcommand{\hps}{HPs\xspace}

\newcommand{\Whh}{\mathbf{W^{hh}}}
\newcommand{\Whx}{\mathbf{W^{hx}}}

\newcommand{\Wyh}{\mathbf{W^{yh}}}
\newcommand{\Wyx}{\mathbf{W^{yx}}}

\newcommand{\Wgyh}{\mathbf{W^{g^yh}}}

\newcommand{\Wghh}{\mathbf{W^{g^hh}}}
\newcommand{\Wgyx}{\mathbf{W^{g^yx}}}
\newcommand{\Wghx}{\mathbf{W^{g^hx}}}

\newcommand{\bby}{\mathbf{b^y}}

\newcommand{\ft}{\text{s}}
\newcommand{\relu}{\text{ReLU}}
\newcommand{\bfg}{b^{fg}}

\newcommand{\mb}{\mathbf}
\newcommand{\mc}{\mathcal}

\usepackage{xcolor}
\definecolor{straightblingin}{rgb}{0.83,0.686,0.216}
\definecolor{darkgreen}{rgb}{0,0.6,0}
\definecolor{darkred}{rgb}{0.7,0.0,0}
\definecolor{orange}{rgb}{1.0, 0.6, 0.0}

\title{Capacity and %
Trainability 
in Recurrent \\ Neural Networks}

\author{Jasmine Collins\thanks{Work done as a member of the Google Brain Residency program (\url{g.co/brainresidency}).}, Jascha Sohl-Dickstein \& David Sussillo \\
Google Brain\\
Google Inc.\\
Mountain View, CA 94043, USA \\
\texttt{\{jlcollins, jaschasd, sussillo\}@google.com} \\
}

\begin{document}

\iclrfinalcopy

\maketitle

\begin{abstract}
Two potential bottlenecks on the expressiveness of recurrent neural networks
(RNNs) are their ability to store information about the task in their
parameters, and to store information about the input history in their
units. We show experimentally that all common RNN architectures achieve nearly
the same per-task and per-unit capacity bounds with careful training, for a variety of tasks and stacking depths. 
They can store an amount of task information which is linear in the number of parameters, and is approximately 5 bits per parameter.
They can additionally store approximately one real
number from their input history per hidden unit. 
We further find that for several tasks it is the per-task
parameter capacity bound that determines performance.  These results suggest
that many previous results comparing RNN architectures are driven primarily by
differences in training effectiveness, rather than differences in capacity.
Supporting this observation, we compare training difficulty for several
architectures, and show that vanilla RNNs are far more difficult to train, yet have slightly higher capacity.
Finally, we propose two novel RNN architectures, 
one of which is easier to train than the LSTM or GRU for deeply stacked architectures.
\end{abstract}

\section{Introduction}
Research and application of recurrent neural networks (RNNs) have seen
explosive growth over the last few years, \citep{martens2011learning, graves2009novel},  and RNNs have become the central component for
some very successful model classes and application domains in deep
learning (speech recognition
\citep{DBLP:journals/corr/AmodeiABCCCCCCD15}, seq2seq
\citep{sutskever2014sequence}, neural machine translation
\citep{bahdanau2014neural}, the DRAW model \citep{gregor2015draw}, educational applications \citep{piech2015deep}, and
scientific discovery \citep{mante2013context}).  Despite these recent
successes, it is widely acknowledged that designing and training the
RNN components in complex models can be extremely tricky.  Painfully 
acquired RNN expertise is still crucial to the success of most
projects.

One of the main strategies involved in the deployment of RNN models is
the use of the Long Short Term Memory (LSTM) networks \citep{hochreiter1997long}, and more recently the
Gated Recurrent Unit (GRU) proposed by \citet{cho2014learning,chung2014empirical} (we refer to these as
gated architectures). The resulting models are perceived as being more easily trained, and achieving lower error.  While it is widely
appreciated that RNNs are universal approximators
\citep{doya1993universality}, an unresolved question is the degree to which gated models are more
computationally powerful in practice, as opposed to simply being easier to train.

Here we provide evidence that the observed superiority of gated models over
vanilla RNN models is almost exclusively driven by trainability.  First we
describe two types of capacity bottlenecks that various RNN architectures
might be expected to suffer from: parameter efficiency related to learning the
task, and the ability to remember input history.  Next, we describe our
experimental setup where we disentangle the effects of these two bottlenecks,
including training with extremely thorough hyperparameter (HP) optimization.  Finally, we
describe our capacity experiment results (per-parameter and per-unit), as well
as the results of trainability experiments (training on extremely hard tasks where
gated models might reasonably be expected to perform better).

\subsection{Capacity Bottlenecks}

There are several potential bottlenecks for RNNs, for example: How much
information about the task can they store in their parameters?  How much
information about the input history can they store in their units?  These
first two bottlenecks can both be seen as memory capacities (one for the task,
one for the inputs), for different types of memory.

Another, different kind of capacity stems from the set of computational
primitives an RNN is able to perform.  For example, maybe one wants to
multiply two numbers.  In terms of number of units and time steps, this task may
be very straight-forward using some specific computational primitives and dynamics, but with others 
it may be extremely resource heavy.  One
might expect that differences in computational capacity due to different computational primitives would play a large role in performance.
However, despite the fact that the gated architectures are outfitted with
a multiplicative primitive between hidden units, while the vanilla RNN is not,
we found no evidence of a computational bottleneck in our experiments.
We therefore will focus only on the per-parameter capacity of an RNN to learn about its
task during training, and on the per-unit memory capacity of an RNN to
remember its inputs.

\subsection{Experimental Setup}

RNNs have many \hps, such as the scalings of matrices and biases,
and the functional form of certain nonlinearities. There are additionally many
\hps involved in training, such as the choice of optimizer, and the
learning rate schedule.
In order to train our models 
we employed a \hp tuner that uses a Gaussian Process model similar
to Spearmint (see Appendix, section on
\hp tuning and
\citet{desautels2014parallelizing,snoek2012practical} for related work).  The
basic idea is that one requests \hp values from the tuner, runs the 
optimization to completion using those values, 
and then returns the validation loss.  This loss is then used by the tuner, in combination
with previously reported losses, to choose new \hp values such
that over many experiments, the validation loss is minimized with respect to
the \hps.  For our experiments, we report the evaluation
loss (separate from the validation loss returned to the \hp optimizer, except where otherwise noted) 
after the \hp tuner has highly optimized the task (hundreds to
many thousands of experiments for each
architecture and task).

In our studies we used a variety of well-known RNN architectures: standard
RNNs such as the vanilla RNN and the newer IRNN \citep{le2015simple}, as well as gated
RNN architectures such as the GRU and LSTM.  We rounded out our set of models by
innovating two novel (to our knowledge) RNN architectures (see Section \ref{rnn architectures}) we call the Update Gate RNN (UGRNN), and the Intersection RNN (+RNN).  The UGRNN is a
`minimally gated' RNN architecture that has only a coupled gate between the
recurrent hidden state, and the update to the hidden state.  The +RNN uses coupled gates to gate both the recurrent 
and depth dimensions in a straightforward way.

To further explore the various strengths and weaknesses of each RNN architecture,
we also used a variety of network depths: 1, 2, 4, 8, in our experiments.\footnote{Not all experiments used a depth of 8, due to limits on computational resources.} In most 
experiments, we held the number of parameters fixed across different architectures and different depths.
More precisely, for a given experiment, a maximum number of parameters was
set, along with an input and output dimension.  The number of
hidden units per layer was then chosen such that the number of parameters, summed across 
all layers of the network, was as large as possible without exceeding the allowed maximum.

For each of our 6 tasks, 6 RNN variants, 4 depths, and 6+ model sizes, we ran the
\hp tuner in order to optimize the relevant loss function.
Typically this resulted in many hundreds to several thousands of
\hp evaluations, each of which was a full training run up to
millions of training steps. Taken together, this amounted to
CPU-millennia worth of computation.

\subsection{Related Work}

While it is well known that RNNs are universal approximators of arbitrary
dynamical systems \citep{doya1993universality}, there is little theoretical
work on the task-capacity of RNNs. \citet{koiran1998vapnik} studied
the VC dimension of RNNs, which provides an upper bound on their
task-capacity (defined in Section \ref{sec ppc}). These upper bounds are not a close match to our experimental results.
For instance, we find that performance saturates rapidly in terms of the number of unrolling steps (Figure \ref{fig more capacity tasks}b), while the relevant bound increases linearly with the number of unrolling steps. "Unrolling" refers to recurrent computation through time.

Empirically, \citet{karpathy2015visualizing}
have studied how LSTMs encode information in character-based text modeling
tasks. Further, \citet{sussillo2013opening} have
reverse-engineered the vanilla RNN trained on simple tasks, using the tools
and language of nonlinear dynamical systems theory. 
In \citet{ISAN} the behavior of switched affine recurrent networks is carefully examined.

The ability of RNNs to store information about their input has been better 
studied, in both the context of machine learning and theoretical neuroscience.
Previous work on short term memory traces explores the tradeoffs between memory fidelity 
and duration, for the case that a new input is presented to the RNN at every
time step \citep{jaeger2004harnessing,maass2002real,white2004short,ganguli2008memory,charles2014short}.
We use a simpler capacity measure consisting only of
the ability of an RNN to store a single input vector. 
Our results suggest that, contrary to common belief, the capacity of
RNNs to remember their input history is not a practical limiting factor on their
performance.

The precise details of what makes an RNN architecture perform well is an
extremely active research field (e.g. \citet{jozefowicz2015empirical}).  A
highly related article is \citet{greff2015lstm}, in which the authors used
random search of \hps, along with systematic removal of pieces of
the LSTM architecture to determine which pieces of the LSTM were more important
than the others. Our UGRNN architecture is directly inspired by the
large impact of removing the forget gate from the LSTM
\citep{gers1999learning}. \citet{Zhou2016} introduced an architecture with minimal gating that is similar to the UGRNN, but is directly inspired by the GRU. An in-depth comparison between RNNs and GRUs in the
context of end-to-end speech recognition and a limited computational budget
was conducted in \citet{DBLP:journals/corr/AmodeiABCCCCCCD15}.  Further, ideas
from RNN architectures that improve ease of training, such as forget gates
\citep{gers1999learning}, and copying recurrent state from one time step to
another, are making their way into deep feed-forward networks as highway
networks \citep{srivastava2015highway} and residual connections \citep{he2015deep}, respectively.  Indeed, the +RNN was inspired in 
part by the coupled depth gate of \citet{srivastava2015highway}.

\subsection{Recurrent Neural Network Architectures} \label{rnn architectures}
Below we briefly define the RNN architectures used in this study.  Unless otherwise stated 
$\WW$ denotes a matrix, $\bb$ denotes a vector of biases.  The symbol $\xxt$ is the 
input at time $t$, and $\hht$ is the hidden state at time $t$.  Remaining vector 
variables represent intermediate values.  The function  $\sigma(\cdot)$ denotes 
the logistic sigmoid function and $\ft(\cdot)$ is either $\tanh$ or $\relu$, set 
as a \hp (see Appendix, Section RNN \hps for the complete 
list of \hps).  Initial conditions for the networks were set to a learned 
bias. Finally, it is a well-known trick of the trade to initialize the gates of an 
LSTM or GRU with a large bias to induce better gradient flow.  We included this 
parameter, denoted as $\bfg$, and tuned it along with all other \hps.

\subsubsection*{RNN, IRNN \citep{le2015simple}}
\begin{align}
  \hht = \ft\left(\Wh \hhtm + \Wx\xxt + \bbh \right)
\end{align}
Note the IRNN is identical in structure to the vanilla RNN, but with an identity initialization
for $\Wh$, zero initialization for the biases, and $\ft = \relu$ only.

\subsubsection*{UGRNN - Update Gate RNN}
Based on \citet{greff2015lstm}, where they noticed the forget gate ``was
crucial'' to LSTM performance, we tried an RNN variant where we began with a
vanilla RNN and {\it added} a single gate. 
This gate determines whether the hidden state is carried over from the previous time step, or updated -- hence, it is an update gate. An alternative way to view the UGRNN is a highway layer gated through time \citep{srivastava2015highway}.
\begin{align}
  \cct &= \ft\left(\Wch\; \hhtm + \Wcx\;\xxt + \bbc \right) \\
  \ggt &= \sigma\left(\Wgh\; \hhtm + \Wgx\;\xxt + \bbg + \bfg \right) \\
  \hht &= \ggt\cdot\hhtm + (\mathbf{1}-\ggt)\cdot\cct
\end{align}

\subsubsection*{GRU - Gated Recurrent Unit \citep{cho2014learning}}
\begin{align}
  \rrt &= \sigma\left(\Wrh\; \hhtm + \Wrx\;\xxt + \bbr \right) \\
  \uut &= \sigma\left(\Wuh\; \hhtm + \Wux\;\xxt + \bbu + \bfg \right) \\
  \cct &= \ft\left(\Wch\; (\rrt \cdot \hhtm) + \Wcx\;\xxt + \bbc \right) \\
  \hht &= \uut\cdot\hhtm + (\mathbf{1}-\uut)\cdot\cct
\end{align}

\subsubsection*{LSTM - Long Short Term Memory\citep{hochreiter1997long}}
\begin{align}
  \iit &= \sigma\left(\Wih\; \hhtm + \Wix\;\xxt + \bbi \right) \\
  \fft &= \sigma\left(\Wfh\; \hhtm + \Wfx\;\xxt + \bbf + \bfg \right) \\
  \mathbf{c}_t^{in} &= \ft\left( \Wch\; \hhtm + \Wcx\;\xxt + \bbc \right) \\
  \cct &= \fft \cdot \mathbf{c}_{t-1} + \iit \cdot \mathbf{c}_t^{in} \\
  \oot &= \sigma\left(\Woh\; \hhtm + \Wox\;\xxt + \bbo  \right) \\
  \hht &= \oot \cdot \tanh(\cct)
\end{align}

\subsubsection*{+RNN - Intersection RNN}
Due to the success of the UGRNN for shallower architectures in this study 
(see later figures on trainability), as well as some of the observed trainability 
problems for both the LSTM and GRU for deeper architectures (e.g. Figure \ref{fig hard tasks}h) we developed the Intersection RNN (denoted with a `+') architecture with a coupled depth gate in addition to a coupled recurrent gate.  Additional influences for this architecture were the recurrent gating of the LSTM and GRU, and the depth gating 
from the highway network \citep{srivastava2015highway}. This architecture has recurrent 
input, $\hhtm$, and depth input, $\xxt$.  It also has recurrent output, $\hht$, and 
depth output, $\yyt$.  Note that this architecture only applies between layers where $\xxt$ 
and $\yyt$ have the same dimension, and is not appropriate for networks 
with a depth of 1 (we exclude depth one +RNNs in our experiments). 
\begin{align}
    \yyt^{in} &= \mbox{s1}\;(\Wyh \; \hhtm + \Wyx\;\xxt + \bby) \\
    \hht^{in} &= \mbox{s2}\;(\Whh \; \hhtm + \Whx\;\xxt + \bbh) \\
    \ggt^y &= \sigma\left(\Wgyh\; \hhtm + \Wgyx\;\xxt + \bbg^y + b^{fg,y} \right) \\
    \ggt^h &= \sigma\left(\Wghh\; \hhtm + \Wghx\;\xxt + \bbg^h + b^{fg,h} \right) \\
    \yyt &= \ggt^{y} \cdot \xxt + (1-\ggt^y) \cdot \mathbf{y}_t^{in} \\
    \hht &= \ggt^{h} \cdot \hhtm + (1-\ggt^h) \cdot \mathbf{h}_t^{in} 
\end{align}
In practice we used $\relu$ for s1 and $\tanh$ for s2.

\section{Capacity Experiments}

\subsection{Per-parameter capacity}\label{sec ppc}
A foundational result in machine learning is that a single-layer
perceptron with $N^2$ parameters can store at least 2 bits of information per
parameter \citep{cover1965geometrical,gardner1988space,baldi1987number}. 
More precisely, a perceptron can implement a mapping from $2N$, $N$-dimensional, 
input vectors to arbitrary $N$-dimensional binary output
vectors, subject only to the extremely weak restriction that the input vectors
be in general position. RNNs provide a far more complex input-output mapping, 
with hidden units, recurrent dynamics, and a diversity of nonlinearities.
Nonetheless, we wondered if there were analogous capacity results for RNNs
that we might be able to observe empirically.

\subsubsection{Experimental Setup}

As we will show in Section \ref{sec additional tasks}, tasks with complex temporal dynamics, such 
as language modeling, exhibit a per-parameter capacity bottleneck that explains the performance of RNNs far better than a per-unit bottleneck.
To make the experimental design as simple as possible, and to remove potential confounds stemming 
from the choice of temporal dynamics, we study per-parameter capacity using a task inspired by \citet{gardner1988space}.
Specifically, to measure how much task-related information can be stored in the parameters
of an RNN, we use a memorization task, where a random static input is injected into an RNN, 
and a random static output is read out some number of time steps later. We emphasize that the same per-parameter 
bottleneck that we find in this simplified task also arises in more temporally complex tasks, such as language modeling.

At a high level, we draw a fixed set
of random inputs and random labels, and train the RNN to map random inputs to randomly 
chosen labels via cross-entropy error.  However, rather than returning the 
cross-entropy error to the \hp tuner (as is normally done), we instead 
return the mutual information between the RNN outputs and the true labels.  In 
this way, we can treat the number of input-output mappings as a \hp, 
and the tuner will select for us the correct number of mappings so as to maximize 
the mutual information between the RNN outputs and the labels.  From this mutual 
information we compute bits per parameter, which provides a normalized measurement of 
how much the RNN learned about the task.

More precisely, we draw datasets of binary inputs $\mb X$ and target binary 
labels $\mb Y$ at uniform from the set of all binary datasets, 
$\mb X \sim \mc X = \left\{0,1\right\}^{n_{in} \times b}$, $\mb Y \sim \mc Y = \left\{0,1\right\}^{1 \times b}$, 
where $b$ is the number of samples, and $n_{in}$ is the dimensionality of the inputs. 
Number of samples, $b$, is treated as a \hp and in practice the optimal dataset size is very close to the bits of mutual information between true and predicted labels. This trend is demonstrated in Figure \ref{fig nm vs mi} in the Appendix.
For each value of $b$ the RNN is trained to minimize the cross entropy of the network output with the true labels.  We write the output of the RNN for all inputs as $\hat{\mb Y} = f\left(\mb X\right)$, 
with corresponding random variable $\hat{\mc Y}$.  
We are interested in the mutual information $I\left( \mc Y; \hat{\mc Y} \right)$ 
between the true class labels and the class labels predicted by the RNN. This 
is the amount of (directly recoverable) information that the RNN has stored 
about the task.  In this setting, it is calculated as 
\begin{align}
I\left( \mc Y; \hat{\mc Y}\right)
& =
H\left( \mc Y \right)
-
H\left( \mc Y | \hat{\mc Y} \right) \label{entropy}\\
& =
b + b\left(
  p \log_2 p + \left(1 - p\right)\log_2 \left(1 - p\right)
\right)
,
\end{align}
where $p$ is the fraction of correctly classified samples.  The number $b$ is then
adjusted, along with all the other \hps, so as to maximize the mutual information 
$I\left( \mc Y; \hat{\mc Y} \right)$.
In practice $p$ is computed using only a single draw of $\{\mb X, \mb Y\} \sim \mc X \times \mc Y$.

We performed this optimization of $I\left( \mc Y; \hat{\mc Y} \right)$ for various RNN architectures, depths, 
and numbers of parameters.  We plot the best value of $I\left( \mc Y;\hat{\mc Y}\right)$ vs.  number
of parameters in Figure \ref{fig capacity tasks}a.  This captures the
amount of information stored in the parameters about the mapping between $\mb
X$ and $\mb Y$.  To get an estimate of bits per parameter, we divide by
the number of parameters, as shown in Figure \ref{fig capacity tasks}e.  

\subsubsection{Results}

\begin{figure*}[h]
\centering

\adjincludegraphics[width=1.0\linewidth]{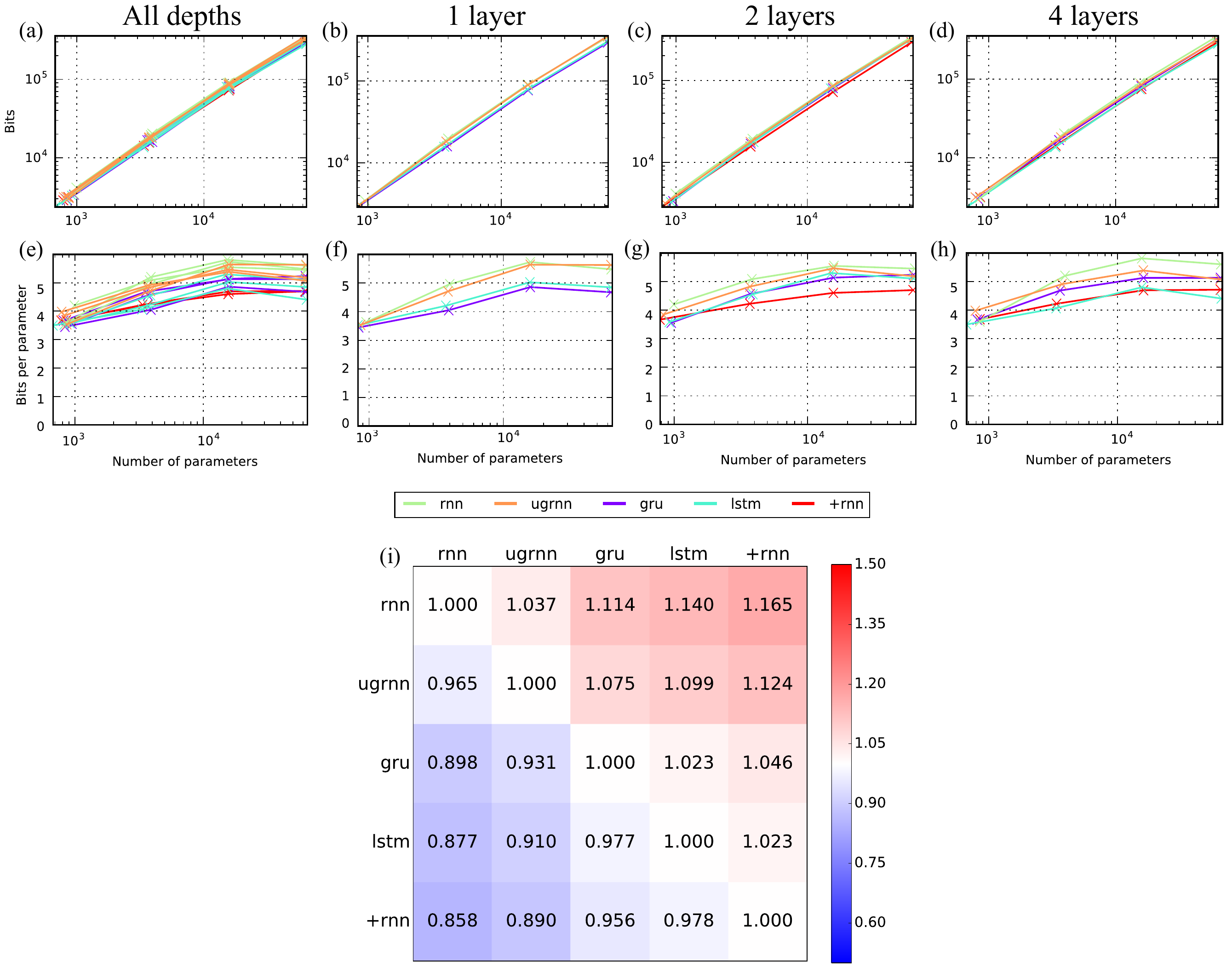}
\caption{
All neural network architectures can store approximately five bits per parameter about a task, with only small variations across architectures.
{\em (a)} Stored bits as a function of network size.
These numbers represent the maximum stored bits across 1000+ \hp optimizations with 5 time steps unrolled at each network size for all levels of depth.
{\em (b-d)} Same as (a), but each level of depth shown separately. 
{\em (e-h)} Same as (a-d) but showing bits per parameter as a function of network size.
{\em (i)} The value in cell $(x,y)$ is the multiplier for the number of parameters needed to give the architecture on the $x$-axis the same capacity as the architecture on the $y$-axis. Capacities are measured by averaging the maximum stored bits per parameter for each architecture across all sizes and levels of depth.}
\label{fig capacity tasks}
\end{figure*}

\begin{figure*}
\centering
\begin{tabular}{cc}
\adjincludegraphics[width=.95\linewidth]{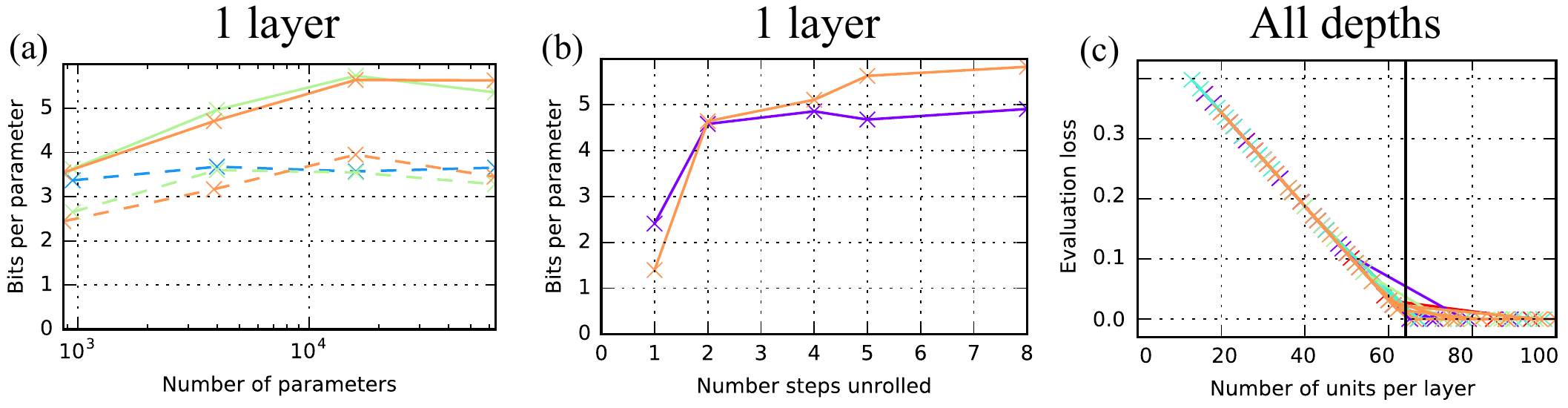} \\
\adjincludegraphics[width=.9\linewidth]{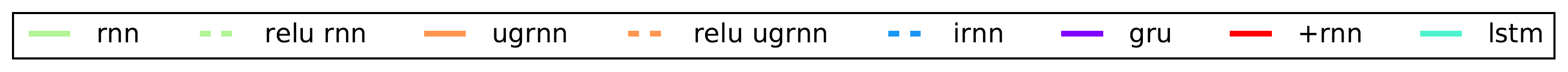}
\end{tabular}
\caption{
Additional RNN capacity analysis. 
{\em (a)} The effect of the $\relu$ nonlinearity on capacity. Solid lines indicate bits per parameter for 1-layer architectures (same as Figure \ref{fig capacity tasks}b), where both $\tanh$ and $\relu$ are nonlinearity choices for the \hp tuner. Dashed lines show the maximum bits per parameter for each architecture when only results achieved by the $\relu$ nonlinearity are considered.
{\em (b)} Bits per parameter as a function of the number of time steps unrolled.
{\em (c)} L2 error curve for all architectures of all depths on the memory throughput task.
The curve shows the error plotted as a function of the number of units for a random input of dimension 64 (black vertical line).
All networks with with less than 64 units have error in reconstruction, while all networks with number of units greater than 64 nearly perfectly reconstruct the random input.
}
\label{fig more capacity tasks}
\end{figure*}

\paragraph{Five Bits per Parameter}
Examining the results of Figure \ref{fig capacity tasks}, we find the capacity of all architectures is roughly linear in the number of parameters, across several orders of magnitude of parameter count. We further find that the capacity is between 3 and 6 bits per parameter,
once again 
across all architectures, depths 1, 2 and 4, and across several orders of magnitude in terms of number of parameters.
Given the
possibility of small size effects, and a larger portion of weights used as
biases at a small number of parameters, we believe our estimates for
larger networks are more reliable. 
This leads us to a bits per parameter 
estimate of approximately 5, averaging over all architectures and all depths. 
Finally, we note that the per-parameter task capacity increases as a function of the 
number of unrollings, though with diminishing gains (Figure \ref{fig more capacity tasks}b).

The finding that our results are consistent across diverse architectures and scales is even more surprising, since prior to these experiments it was not clear that capacity would even scale linearly with the number of parameters. For instance, previous results on model compression -- by reducing the number of parameters \citep{yang2015deep}, or by reducing the bit depth of parameters \citep{hubara2016binarized} -- might lead one to predict that different architectures use parameters with vastly different efficiencies, and that task capacity increases only sublinearly with parameter count.

\paragraph{Gating Slightly Reduces Capacity}
While overall, the different architectures performed very similarly, there are some 
capacity differences between architectures that appear to hold up across most depths 
and parameter counts.  To quantify these differences we constructed a table showing the change in the number of parameters one would need to switch from one architecture to another, while maintaining equivalent capacity (Figure \ref{fig capacity tasks}i).  One trend that emerged from our capacity experiments is a slightly 
reduced capacity as a function of "gatedness".  Putting aside the IRNN, which performed 
the worst and is discussed below, we noticed that across all depths and all model sizes, 
the performance was on average RNN > UGRNN > GRU > LSTM > +RNN.  The vanilla RNN 
has no gates, the UGRNN has one, while the remaining three have two or more.

\paragraph{ReLUs Reduce Capacity}
In our capacity tasks, the IRNN performed noticeably worse than all other architectures, reaching a maximum bits per parameter of roughly 3.5. To determine if this performance drop was due to the $\relu$ nonlinearity of the IRNN, or its identity initialization, we sorted through the RNN and UGRNN results (which both have $\relu$ and $\tanh$ as choices for the nonlinearity \hp) and looked at the maximum bits per parameter when only optimizations using $\relu$ are considered. Indeed, both the RNN and UGRNN bits per parameter dropped dramatically to the 3.5 range (Figure \ref{fig more capacity tasks}a) when those architectures exclusively used $\relu$, providing strong evidence that the $\relu$ activation function is problematic for this capacity task. 

\subsection{Per-unit capacity to remember inputs}

An additional capacity bottleneck in RNNs is their ability to store information about 
their inputs over time.  It may be plainly obvious that an IRNN, which is essentially 
an integrator, can achieve perfect memory of its inputs if the number of inputs is less 
than or equal to the number of hidden units, but it is not so clear for some of the more 
complex architectures.  So we measured the per-unit input memory empirically.  Figure 
\ref{fig more capacity tasks}c shows the intuitive result that every RNN 
architecture (at every depth and number of parameters) we studied can reconstruct a 
random $n_{in}$ dimensional input at some time 
in the future, if and only if the number of hidden units per layer in the network, $n_{h}$, is 
greater than or equal to $n_{in}$ 
Moreover, regardless of RNN architecture, the error 
in reconstructing the input follows the same curve as a function of the number of hidden 
units for all RNN variants, corresponding to reconstructing an $n_{h}$ dimensional 
subspace of the $n_{in}$ dimensional input.

\begin{figure*}[h]
\centering
\begin{tabular}{cc}

\adjincludegraphics[width=1.0\linewidth]{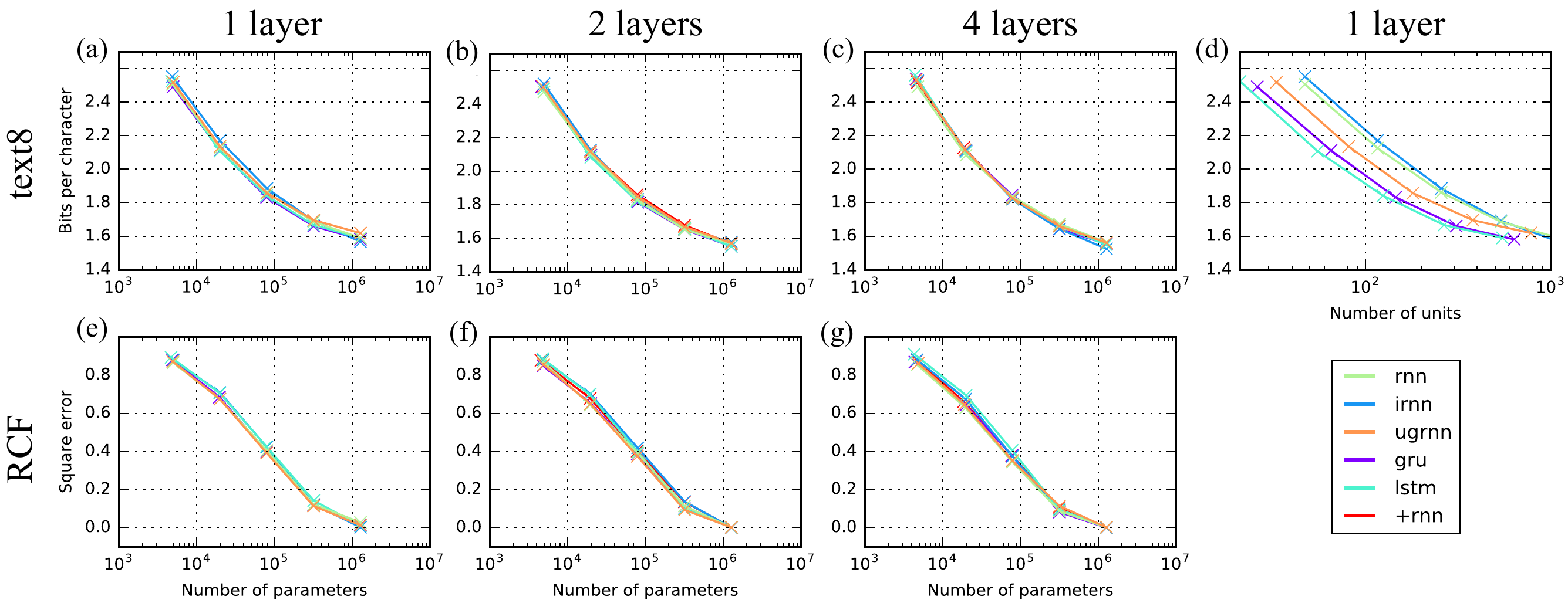}

\end{tabular}
\caption{
All RNN architectures achieved near identical performance given the same number of parameters, on a language modeling and random function fitting task.
{\em (a-c)} text8 Wikipedia number of parameters vs bits per character for all RNN architectures. From left to right: 1 layer, 2 layer, 4 layer models.
{\em (d)} text8 number of hidden units vs bits per character for 1 layer architectures.  We note that this is almost always a misleading way to compare architectures as the more heavily gated architectures appear to do better when compared per-unit.
{\em (e-g)} Same as (a-c), except showing square error for different model sizes trained on RCFs.
}
\label{fig additional tasks}
\end{figure*}

We highlight this per-unit capacity to make the point that a per-parameter task capacity appears 
to be the limiting factor in our experiments (e.g. Figure \ref{fig capacity tasks} and Figure \ref{fig additional tasks}), and not a 
per-unit capacity, such as the per-unit capacity to remember previous inputs.  Thus when 
comparing results between architectures, one should normalize different architectures by the 
number of parameters, and not the number of units, as is frequently done in the literature 
(e.g. when comparing vanilla RNNs to LSTMs).  
This makes further sense as, for all common RNN architectures, the computational cost of processing a single sample is linear in the number of parameters, and quadratic in the number of units per layer.  
As we show in Figure \ref{fig additional tasks}d, plotting the capacity results by numbers of units gives very misleading results.

\section{Additional tasks where architectures achieve very similar loss}
\label{sec additional tasks}
 
We studied additional tasks that we believed to be easy enough to train
that the evaluation loss of different architectures would reveal variations in
capacity rather than trainability.  A critical aspect of these tasks is that
they could not be learned perfectly by any of the model sizes in our
experiments.  As we change model size, we therefore expect performance on the
task to also change.  The tasks are (see Appendix, section Task Definitions
for further elaboration of these tasks):
\begin{itemize} 
  \item text8 - 1-step ahead character-based prediction on the text8 Wikipedia dataset (100 million characters) \citep{text8}.
  \item Random Continuous Functions (RCF) - A task similar to the per-parameter capacity task above, except the
    target outputs are real numbers (not categorical), and the number of training
    samples is held fixed.
\end{itemize}

The performance on these two tasks is shown in Figure \ref{fig additional
  tasks}.  The evaluation loss as a function of the number of parameters is
plotted in panels a-c and e-g, 
for the text8 task, and RCF task,
respectively.  
For all tasks in this section, the number of parameters rather than
the number of units provided the bottleneck on performance, and all
architectures performed extremely closely for the same number of
parameters. By close performance we mean that, for one model to achieve the
same loss as another the model, the number of parameters would have to be
adjusted by only a small factor (exemplified in Figure \ref{fig capacity tasks}i for the per-parameter capacity task).

\section{Tasks that are very hard to learn}

In practice it is widely appreciated that there is often a significant gap in performance
between, for example, the LSTM and the vanilla RNN, with the LSTM nearly always 
outperforming the vanilla RNN.  Our per-parameter capacity results provide evidence for a rough equivalence
among a variety of RNN architectures, with slightly higher capacity in the vanilla RNN (Figure 
\ref{fig capacity tasks}). To reconcile our per-parameter capacity results with widely held experience, we 
provide evidence that gated architectures, such as the LSTM, are far easier to train
than the vanilla RNN (and often the IRNN).

\begin{figure*}[h!]
\centering
\adjincludegraphics[width=1.0\linewidth]{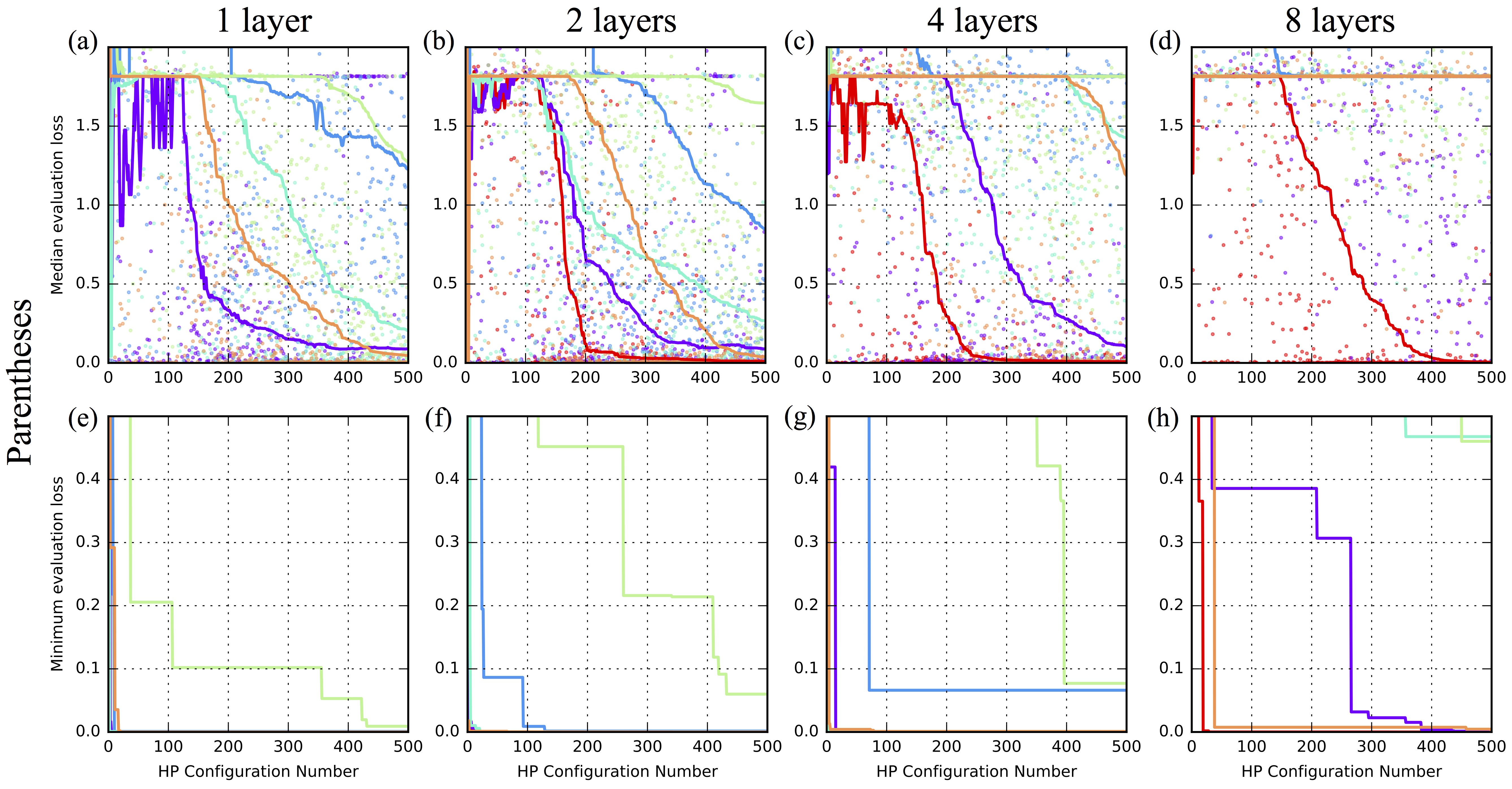}
\adjincludegraphics[width=1.0\linewidth]{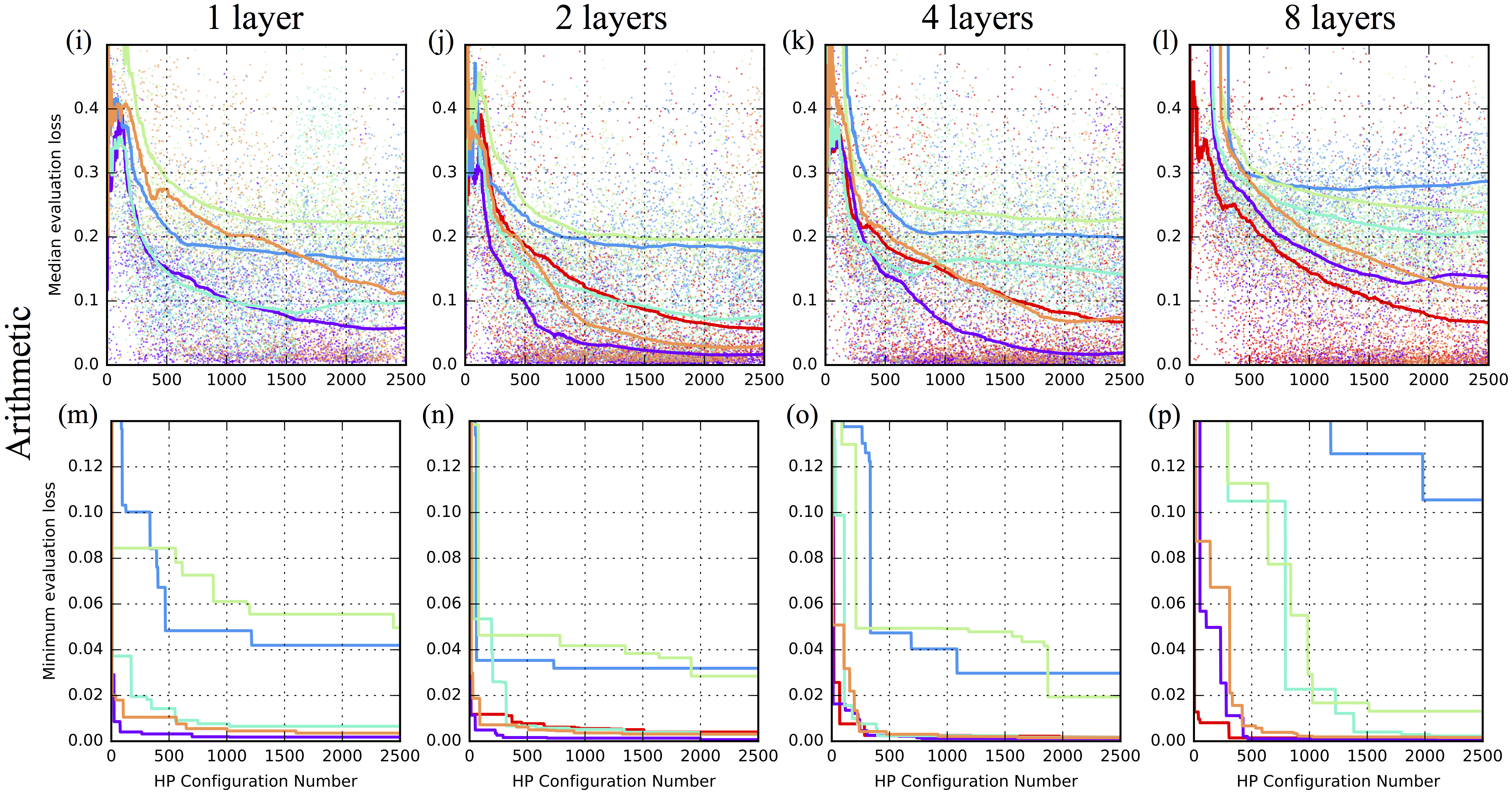}
\adjincludegraphics[width=.7\linewidth]{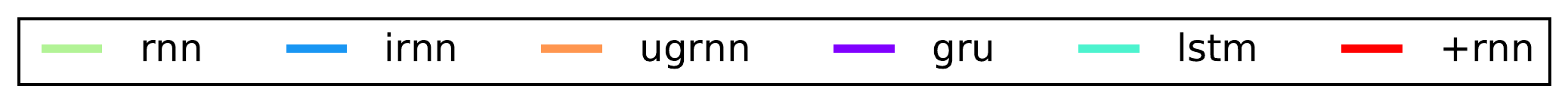}

\caption{ 
Some RNN architectures are far easier to train than others. 
Results of \hp searches on extremely difficult tasks.
{\em (a)} Median evaluation error as a function of \hp optimization iteration for 1 layer architectures on the parentheses task. Dots indicate evaluation loss achieved on that \hp iteration.
{\em (b-d)} Same as (a), but for 2, 4 and 8 layer architectures.
{\em (e-h)} Minimum evaluation error as a function of \hp optimization iteration for parentheses task. Same depth order as (a-d).
{\em (i-p)} Same as (a-h), except for the arithmetic task. We note that the best loss for the vanilla RNN is still decreasing after 2400+ \hp evaluations. 
}
\label{fig hard tasks}
\end{figure*}

We study two tasks that are difficult to learn: parallel parentheses
counting of independent input streams, and mathematical addition of integers 
encoded in a character string (see Appendix, section Task Definitions). The parentheses 
task is moderately difficult
to learn, while the arithmetic task is quite hard. The results of the \hp optimizations are shown in Figure
\ref{fig hard tasks}a-\ref{fig hard tasks}h for the parentheses task, and in Figure 
\ref{fig hard tasks}i-\ref{fig hard tasks}p for the arithmetic task.  These tasks show that, while it is
possible for a vanilla RNN to learn these tasks reasonably well, it is far
more difficult than for a gated architecture.  Note that the best achieved loss on the arithmetic 
task is still significantly decreasing, even after 2500 \hp evaluations (2500 full complete 
optimizations over the training set), for the RNN and IRNN.  

There are three noteworthy trends in these trainability experiments.  First, across both tasks, and 
all depths (1, 2, 4 and 8), the RNN and IRNN performed most poorly, and took the longest
to learn the task.  Note, however that both the RNN and IRNN always solved the tasks eventually,
at least for depth 1.  Second, as the stacking depth increased, the gated architectures 
became the only architectures that could solve the tasks.  Third, %
the most trainable architecture for depth 1 was the GRU, and the most trainable architecture for
depth 8 was the +RNN (which performed the best on both of our metrics for trainability, on both tasks).

\begin{figure*}[h!]
\centering
\begin{tabular}{cc}
\adjincludegraphics[width=0.7\linewidth]{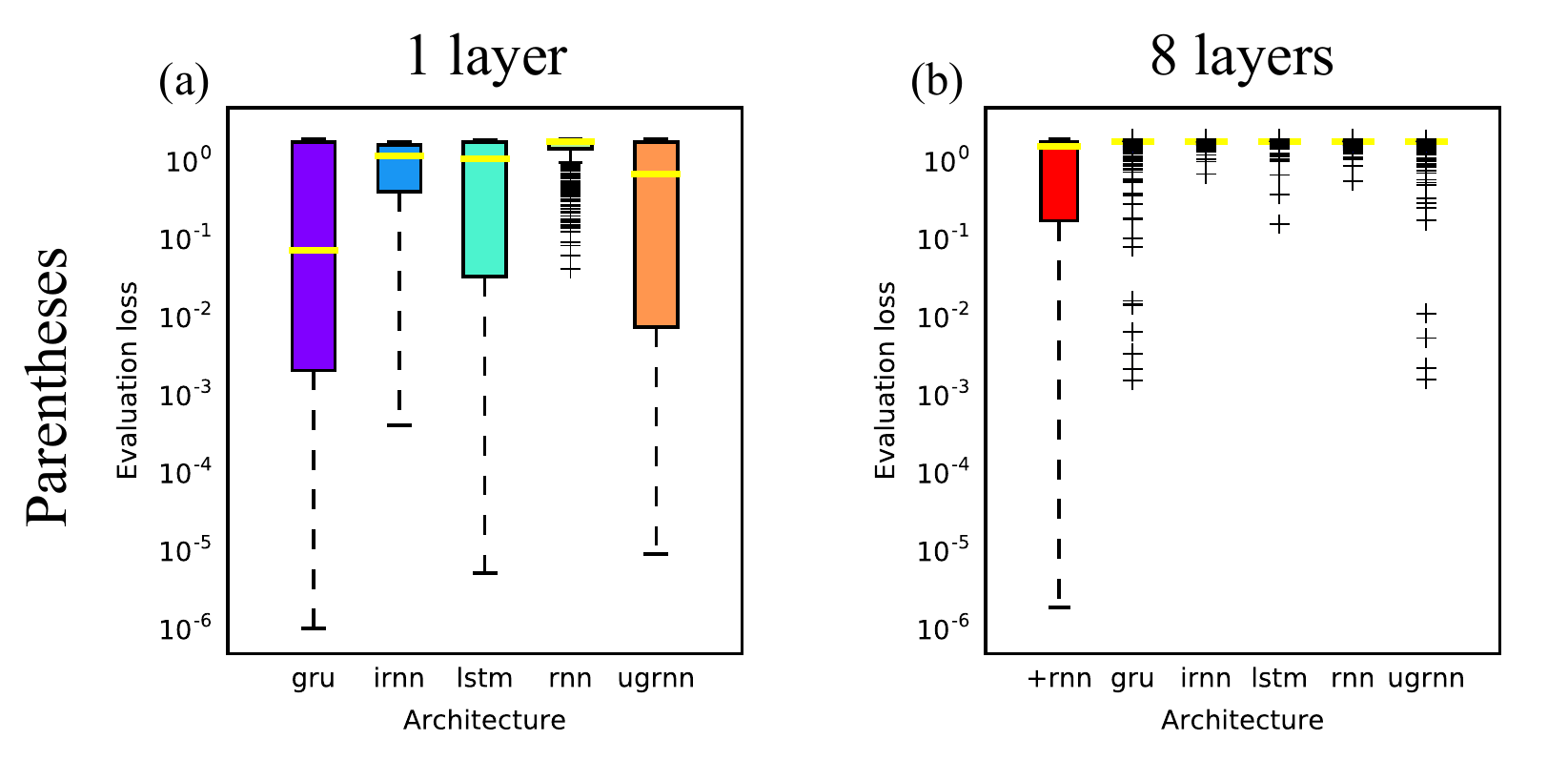}
\end{tabular}
\caption{
For randomly generated hyperparameters, GRU and +RNN are the most easily trainable architectures. 
Evaluation losses from 1000 iterations of randomly chosen \hp sets for 1 and 8 layer, 200k parameter models on the parentheses task. Statistics from a Welch's $t$-test for equality of means on all pairs of architectures are presented in Table \ref{table t test}.
{\em (a)} Box and whisker plot of evaluation losses for the 1 layer model. 
{\em (b)} Same as (a) but for 8 layers. 
}
\label{fig rand}
\end{figure*}

To achieve our results on capacity and trainability, we relied heavily on a \hp tuner.  
Most practitioners do not have the time or resources to make use of such a tuner, typically only adjusting 
the \hps a few times themselves. So we wondered how the various architectures would perform
if we set \hps randomly, within the ranges specified (see Appendix for ranges).  We tried this 1000 times
on the parentheses task, for all 200k parameter architectures at depths 1 and 8 (Figure \ref{fig rand} and Table \ref{table rand infeas}).  
The noticeable trends are that the IRNN returned an infeasible error nearly half of the time, and the LSTM (depth 1) and GRU (depth 8) were infeasible the least number of times, where infeasibility means that the training loss diverged. For depth 1, the GRU gave the smallest error, and the smallest median error,
and for depth 8, the +RNN delivered the smallest error and smallest median error.

\begin{table}[!htb]
\centering
\begin{tabular}{|c|c|c|}
    \hline
    Architecture & \% Infeasible (1 layer) & \% Infeasible (8 layer) \\
    \hline
    +RNN & - & 8.8 \% \\
    GRU & 15.5 \% & 3.2 \% \\
    IRNN & 56.7 \% & 44.6 \% \\
    LSTM & 12.0 \% & 4.0 \% \\
    RNN & 21.5 \% & 18.7 \% \\
    UGRNN & 20.2 \% & 11.5 \% \\
    \hline
\end{tabular}
\caption{
Fraction infeasible trials as a result of 1000 iterations of randomly chosen \hp sets for 1 and 8 layer, 200k parameter models trained on the parentheses task.
} 
\label{table rand infeas}
\end{table}

\section{Discussion}

Here we report that a number of RNN variants can hold between 3-6 bits per
parameter about their task, and that these variants can remember a number of random inputs that is nearly equal to the number
of hidden units in the RNN. 
The quantification of the number of bits per parameter an RNN can store about a task is particularly important, as it was not previously known whether the amount of information about a task that could be stored was even linear in the number of parameters.

While our results point to empirical capacity limits for both task
memorization, and input memorization, apparently the requirement to remember
features of the input through time is not a practical bottleneck.  If it were, then
the vanilla RNN and IRNN would perform better than the gated
architectures in proportion to the ratio of the number of units, which they do not.
Based on widespread results in the literature, and our own results on our difficult tasks, the loss of some
memory capacity (and possibly a small amount of per-parameter storage capacity)
for improved trainability seems a worthwhile trade off. 
Indeed, the input memory capacity did not obviously impact any task not explicitly designed to measure it, 
as the error curves --  for instance for the language modeling task -- overlapped across architectures for the same number of parameters, but not the same number of units.

Our result on per-parameter task capacity, about 5 bits per parameter averaged over 
architectures, is in surprising agreement with
recently published results on the capacity of synapses in biological neurons.
This number was recently calculated to be about 4.7 bits per synapse, based on biological synapses in
the hippocampus having roughly 26 measurable discrete sizes
\citep{bartol2016nanoconnectomic}.  Our capacity results  
have implications for compressed networks that employ quantization techniques. In particular,
they provide an estimate of the number of bits which a weight may be
compressed without loss in task performance. Coincidentally, in \citet{han2015deep}, the
authors used 5 bits per weight in the fully connected layers.

An additional observation about per-parameter task capacity in our experiments
is that it increases for a few time steps beyond one (Figure \ref{fig more capacity tasks}b), and
then appears to saturate.  We interpret this to suggest that recurrence endows
additional capacity to a network with shared parameters, but that there are
diminishing returns, and the total capacity remains bounded even as the number
of time steps increases.

We also note that performance is nearly constant across RNN architectures if the number of parameters is held fixed. This may motivate the design and use of architectures with small compute per parameter ratios, such as mixture of experts RNNs \citep{DBLP:journals/corr/ShazeerMMDLHD17}, and RNNs with large embedding dictionaries on input and output \citep{DBLP:journals/corr/JozefowiczVSSW16}.

Despite our best efforts, we cannot claim that we perfectly trained
any of the models. Potential problems in \hp optimization could be local minima, as well as 
stochastic behavior in the \hp optimization as a result of the stochasticity of batching 
or random draws for weight matrices. We tried to uncover these effects by running 
the best performing \hps 100 times, and did not observe any serious
deviations from the best results (see Table \ref{table best hps} in Appendix). 
Another form of validation comes from the fact that in our capacity task, essentially 3 independent experiments (one for each level of depth) yielded a clustering by architecture (Figure \ref{fig capacity tasks}e).

Do our results yield a framework for choosing a recurrent architecture? In total, we believe yes.
As explored in \citet{DBLP:journals/corr/AmodeiABCCCCCCD15}, a practical concern for recurrent models is speed of execution in
a production environment.  Our results suggest that if one has a large
resource budget for training and confined resource budget for
inference, one should choose the vanilla RNN.  Conversely, if the training resource budget is small, but the
inference budget large, one should choose a gated model.  Another serious concern relates to task complexity. 
If the task is easy to learn, a vanilla RNN should yield good results.  However if the task
is even moderately difficult to learn, a gated architecture is the right choice.  Our results
point to the GRU as being the most learnable of gated RNNs for shallow architectures, followed by the UGRNN.
The +RNN typically performed best for deeper architectures.  Our results on 
trainability confirm the widely held view that the LSTM is an extremely reliable architecture, but 
it was almost never the best performer in our experiments. Of course further experiments will be required to fully vet the UGRNN and +RNN.
All things considered, in an uncertain training environment, our results suggest using the GRU or +RNN.

\section{Acknowledgements}

We would like to thank Geoffrey Irving, Alex Alemi, Quoc Le, Navdeep Jaitly, and Taco Cohen for helpful feedback.

\small
\bibliographystyle{iclr2017_conference}
\bibliography{cap_n_learn_rnn}

\clearpage
\normalsize
\appendix
\part*{Appendix}

\setcounter{figure}{0} \renewcommand{\thefigure}{App.\arabic{figure}}
\setcounter{table}{0} \renewcommand{\thetable}{App.\arabic{table}}

\section{RNN \hps Set by the \hp Tuner} \label{hptuner}

We used a \hp tuner that uses a Gaussian Process (GP) Bandits approach
for \hp optimization.  Our setting of the tuner's internal
parameters was such that it uses Batched GP Bandits with an expected
improvement acquisition function and a Matern 5/2 Kernel with feature scaling
and automatic relevance determination performed by optimizing over kernel
\hps. Please see \citet{desautels2014parallelizing} and
\citet{snoek2012practical} for closely related work.

For all our tasks, we requested \hps from the tuner, and reported
loss on a validation dataset. 
For the per-parameter capacity task, the evaluation, validation and training datasets were identical. For text8, the validation and evaluation sets consisted of different sections of held out data. For all other tasks, evaluation, validation, and training sets were randomly drawn from the same distribution.
The performance we plot in all cases is on the evaluation dataset.

Below is the list of all tunable \hps that were generically applied to all models.
In total, each RNN variant had between 10 and 27 \hp dimensions relating to 
the architecture, optimization, and regularization.
\begin{itemize}
\item $\ft()$ - as used in the following RNN definitions, a nonlinearity determined
  by the \hp tuner, $\{\relu$, $\tanh\}$. The only exception was the IRNN, 
  which used $\relu$ exclusively.
\item For any matrix that is inherently square, e.g. $\Wh$, there were three
  possible initializations: identity, orthogonal, or random normal
  distribution scaled by $1/\sqrt{n_h}$, with $n_h$ the number of recurrent
  units.  The sole exception was the RNN, which was limited to either orthogonal
  or random normal initializations, to differentiate it from the IRNN.  For any 
  matrix that is inherently rectangular, e.g. $\Wx$, we
  initialized with a random normal distribution scaled by $1/\sqrt{n_{in}}$,
  with $n_{in}$ the number of inputs.
\item For all matrix initializations except the identity initialization,
  there was a multiplicative scalar used to set the scale of matrix. The
  scalar was exponentially distributed in $[0.01, 2.0]$ for recurrent matrices
  and $[0.001,2.0]$ for rectangular matrices.  
\item Biases could have two possible distributions: all biases set to a
  constant value, or drawn from a standard normal distribution.
\item For all bias initializations, a multiplicative scalar was drawn,
  uniformly distributed in $[-2.0, 2.0]$ and applied to bias initialization.
\item We included a scalar bias \hp $\bfg$ for architectures that
  contain forget or update gates, as is commonly employed in practice, which
  was uniformly distributed in $[0.0, 6.0]$.
\end{itemize}

Additionally, the \hp tuner was used to optimize \hps
associated with learning:
\begin{itemize}
\item The number of training steps - The exact range varied between tasks, but
  always fell between 50K and 20M.
\item One of four optimization algorithms could be chosen: vanilla SGD, SGD
  with momentum, RMSProp \citep{tieleman2012rmsprop}, or ADAM
  \citep{DBLP:journals/corr/KingmaB14}.
\item learning rate initial value, exponentially distributed in $[\num{1e-4},
  \num{1e-1}]$
\item learning rate decay - exponentially distributed in $[\num{1e-3},
  1]$. The learning rate exponentially decays by this factor over the number of training steps chosen
  by the tuner
\item optimizer momentum-like parameter - expressed as a logit, and uniformly
  distributed in $[1.0, 7.0]$
\item gradient clipping value - exponentially distributed in $[1, 100]$
\item l2 decay - exponentially distributed in $[\num{1e-8}, \num{1e-3}]$.
\end{itemize}

The perceptron capacity task also had associated \hps:
\begin{itemize}
\item The number of samples in the dataset, $b$ - between 0.1x and 10x the number of model parameters
\item A \hp determined whether the input vector $\mb X_{\cdot j}$ was
presented to the RNN only at the first time step, or whether it was presented
at every time step.
\end{itemize}

Some optimization algorithms had additional parameters such as ADAM's
  second order decay rate, or epsilon parameter. These were set to their
  default values and not optimized. The batch size was set individually by hand for all experiments.
The same seed was used to initialize the random number generator for all task
parameters, whereas the generator was randomly seeded for network parameters (e.g. initializations). Note that for each network, the initial condition was set to a learned vector. 

\begin{figure*}[h!]
\centering
\adjincludegraphics[width=0.4\linewidth]{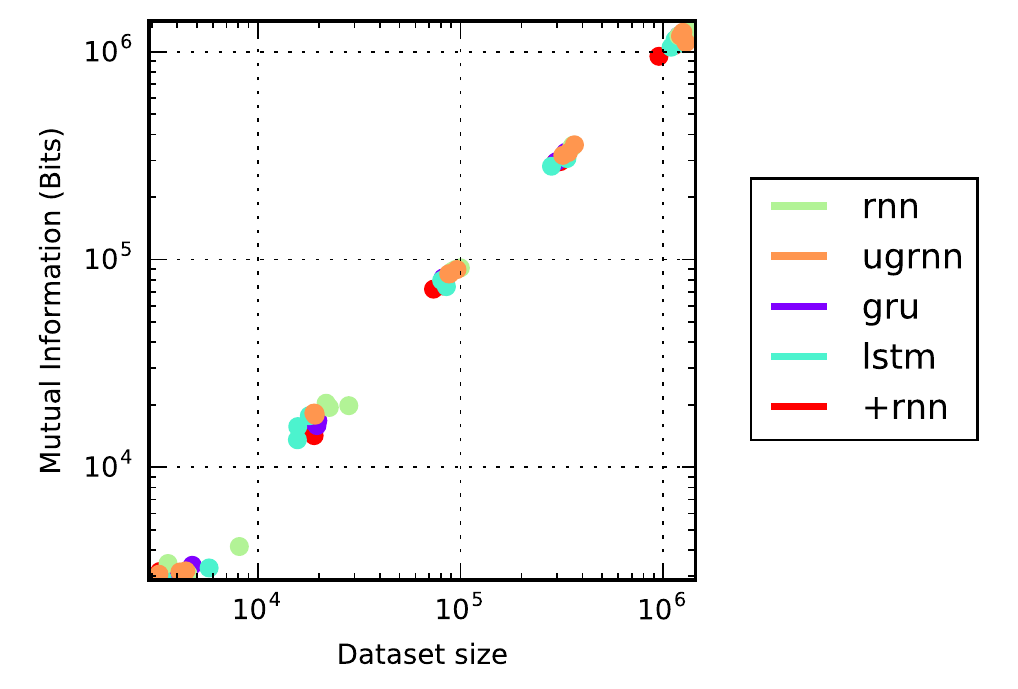} 
\caption{
In the capacity task, the optimal dataset size found by the \hp tuner was only slightly larger than the mutual information in bits reported in Figure \ref{fig capacity tasks}a, for all architectures at all sizes and depths.
}
\label{fig nm vs mi}
\end{figure*}

\section{Task Definitions}   \label{task definitions}

\subsection*{Perceptron Capacity}

While at a high-level, for the perceptron capacity task, we wanted to optimize
the amount of information the RNN carried about true random labels, in
practice, the training objective was standard cross-entropy.  However, when returning
a validation loss to the \hp tuner, we returned the mutual
information $I\left( \mb Y; \hat{\mb Y}|\mb X\right)$.
Conceptually, this is as if there is one
nested optimization inside another.  The inner loop optimizes the RNN for the
set of \hps, training cross entropy, but returning mutual
information.  The outer loop then chooses the \hps, in particular,
the number of samples $b$, in equation (\ref{entropy}), so as to maximize the amount
of mutual information.  This implementation is necessitated because
there is no straightforward way to differentiate mutual information with respect
to number of samples. During training, cross entropy error is evaluated beginning after 5 time steps.

\subsection*{Memory Capacity}
In the Memory Capacity task, we wanted to know how much information an RNN can
reconstruct about its inputs at some later time point.  We picked an input
dimension, 64, and varied the number of parameters in the networks such that
the number of hidden units was roughly centered around 64.  After 12 time steps
the target of the network was exact reconstruction of the input, with a square error loss. The inputs were random values drawn from a uniform distribution between $-\sqrt{3}$ and $\sqrt{3}$ (corresponding to a variance of 1).

\subsection*{Random Continuous Function}
A dataset was constructed consisting of $N=10^6$ random unit norm Gaussian input vectors $\mb x$,
with
size $d=50$. Target scalar outputs $y$ were generated for each input vector, and were
also drawn from a unit norm Gaussian.
Each sample $i$ was assigned a power law weighting $\beta_i = \frac{\left( i + \tau \right)^{-1}}{Z}$,
where $Z$ was a normalization constant such that the weightings summed to 1, and the characteristic
time constant $\tau=5000$.
The loss function for training was calculated after 50 time steps and was weighted square error on the $y_i$,
with the $\beta_i$ acting as the weighting terms.

\subsection*{text8}
In the text8 task, the task was to predict one character ahead in the text8
dataset (1e8 characters of
Wikipedia) \citep{text8}.  Input was a
hot-one encoded sequence, as was the output.  The loss was cross-entropy loss
on a softmax output layer.  Rather than use partial unrolling as is common in
language modeling, we generated random pointers into the text.  The first 13 time
steps (where $T=50$) were used to initialize the RNN into a normal operating mode, and remaining steps were used for training
or inference.

\subsection*{Parentheses Counting Task}
The parentheses counting task independently counts the number of opened
`parens', e.g. `(', without the closing `)'.  Here parens is used to mean any
of 10 parens type pairs, e.g. `<>' or `[]'.  Additionally, there were 10 noise
characters, `a' to `j'.  For each paren type, there was a $20D+10D=30D$
hot-one encoding of all paren and noise symbols, for a total of 300 inputs.
The output for each paren type was a hot-one encoding of the digits 0-9, which
represented the count of the opened parens of that type.  If the count
exceeded 9, the the network kept the count at 9, if the paren was closed, the
count decreased.  The loss was the sum of cross-entropy losses, one for each
paren type.  Finally, for each paren input stream, 50\% random noise
characters were drawn, and 50\% random paren characters were drawn, e.g. 10
streams like `(a{<a<bcb>[[[)'.  Parens of other types were treated as noise for
        the current type, e.g. for the above string if the paren type was
        `<>', the answer is `1' at the end.  The loss was defined only at the
        final time point, $T$, and $T=175$.

\subsection*{Arithmetic Task}
In the arithmetic task, a hot-one encoded character sequence of an addition
problem was presented as input to the network, e.g., `-343243+93851= ', and
the output was the hot-one encoded answer, including the correct amount of
left padded spaces, `-249392'.  An additional \hp for this task was
the number of compute steps (1-6) between the input of the `=' and the first
non-space character in the target output sequence.  The two numbers in the
input were randomly, uniformly selected in $[\num{-1e7},\num{1e7}]$. After 36 time steps, cross-entropy loss was calculated. We found
this task to be extremely difficult for the networks to learn, but when the
task was learned, certain of the network architectures could perform the task
nearly perfectly.

\section{\hp Robustness}   \label{hp robustness}
We wondered how robust the \hps are to the variability of both random batching of data,
and random initialization of parameters.  So we identified the best \hps from the 
parentheses experiments of 100k parameter, 1 layer architectures, and reran the 
parameter optimization 100 times.  We measured the number of infeasible experiments, as well as a 
number of statistics of the loss for the reruns (Table \ref{table best hps}).  These results show that the 
best \hps yielded a distribution of losses very close to the originally reported loss value.
\begin{table}[h!]
\centering
\begin{tabular}{|c c | c c c c c c|} 
 \hline
 Architecture & Original & Infeasible & Min & Mean & Max & S.D. & S.D./Mean \\ [0.75ex] 
 \hline
 RNN & 1.16e-2 & 0 \% & 1.41e-2 & 8.21e-2 & 0.294 & 5.22e-2 & 0.636  \\ 
 IRNN & 4.20e-4 & 48 \% & 2.24e-4 & 5.02e-4 & 8.69e-4 & 1.35e-4 & 0.269 \\
 UGRNN & 1.02e-4 & 0 \% & 3.66e-5 & 2.71e-4 & 6.06e-3 & 7.12e-4 & 2.63 \\
 GRU & 2.80e-4 & 1 \% & 7.66e-5 & 1.89e-4 & 5.48e-4 & 9.08e-5 & 0.480  \\
 LSTM & 7.96e-4 & 0 \% & 8.10e-4 & 2.02e-3 & 0.0145 & 2.31e-3 & 1.14 \\ [1ex] 
 \hline
\end{tabular}
\caption{
Results of 100 runs on the parentheses task using the best \hps for each architecture, at depth 1. \hps were chosen to be the set which achieved the minimum loss. Table shows original loss achieved by the \hp tuner, amount of infeasible trials, minimum loss from running 100 iterations of the same \hps, mean loss, maximum loss, standard deviation, and standard deviation divided by the mean.}
\label{table best hps}
\end{table}

\begin{table}[h!]
\centering
\begin{tabular}{|c | c c c | c c c |} 
 \cline{2-7}
 \multicolumn{1}{c}{ } &
 \multicolumn{3}{|c|}{1 layer} & 
 \multicolumn{3}{c|}{8 layer} \\
 \hline
 & $t$-stat & df & $p$-value  & $t$-stat & df & $p$-value  \\ [0.75ex] 
 \hline
+RNN/GRU & - & - & - & -23.6 & 1080 & < 0.001 \\ 
+RNN/IRNN & - & - & - & -25.7 & 954 & < 0.001 \\ 
+RNN/LSTM & - & - & - & -26.1 & 941 & < 0.001 \\ 
+RNN/RNN & - & - & - & -25.8 & 946 & < 0.001 \\ 
+RNN/UGRNN & - & - & - & -24.3 & 1050 & < 0.001 \\ 
GRU/IRNN & -7.74 & 696 & < 0.001 & -3.51 & 1360 & < 0.001 \\ 
GRU/LSTM & -6.65 & 1750 & < 0.001 & -4.84 & 1290 & < 0.001  \\
GRU/RNN & -26.5 & 1340 & < 0.001 & -3.93 & 1330 & < 0.001  \\
GRU/UGRNN & -4.11 & 1620 & < 0.001 & -1.13 & 1840 & \textbf{0.261}  \\
IRNN/LSTM & 2.23 & 652 & 0.0264 & -2.04 & 1250 & 0.0420  \\
IRNN/RNN & -12.7 & 426 & < 0.001 & -0.571 & 1250 & \textbf{0.568}  \\
IRNN/UGRNN & 4.03 & 719 & < 0.001 & 2.37 & 1320 & 0.0178  \\
LSTM/RNN & -19.6 & 1500 & < 0.001 & 1.53 & 1730 & \textbf{0.125}  \\
LSTM/UGRNN & 2.25 & 1640 & 0.0247 & 3.81 & 1260 & < 0.001  \\
RNN/UGRNN & 20.7 & 1210 & < 0.001 & 2.81 & 1300 & 0.00498  \\ [1ex] 
 \hline
\end{tabular}
\caption{Results of Welch's $t$-test for equality of means on evaluation losses of architecture pairs trained on the parentheses task with randomly sampled \hps. 8 layer GRU and UGRNN, IRNN and RNN, and LSTM and RNN pairs have loss distributions that are different with statistical significance ($p$ > 0.05). Negative $t$-statistic indicates that the mean of the second architecture in the pair is larger than the first.}
\label{table t test}
\end{table}

\end{document}